\documentclass[letterpaper]{article} 
\usepackage{aaai2027-arxiv}
\usepackage[hyphens]{url}  
\usepackage{graphicx} 
\usepackage{natbib}  
\usepackage{caption} 
\usepackage{algorithm}
\usepackage{algorithmic}

\usepackage{newfloat}
\usepackage{listings}
\DeclareCaptionStyle{ruled}{labelfont=normalfont,labelsep=colon,strut=off} 
\floatstyle{ruled}
\newfloat{listing}{tb}{lst}{}
\floatname{listing}{Listing}

\usepackage{booktabs}
\usepackage{colortbl}
\usepackage{pifont}
\newcommand{\cmark}{\ding{51}}
\newcommand{\xmark}{\ding{55}}
\usepackage{subfiles}
\usepackage{xcolor}
\definecolor{niceblue}{RGB}{65,130,240}
\definecolor{nicered}{RGB}{220,50,47}
\usepackage[colorlinks=true, citecolor=niceblue, linkcolor=nicered, urlcolor=blue]{hyperref}

\nocopyright 

\title{Faster-WAM: Efficient Inference-Time Future Conditioning \\ for Robust World Action Models}
\author{
    Weiheng Zhao\textsuperscript{\rm 1},
    Haoyi Jiang\textsuperscript{\rm 1},
    Xin Shi\textsuperscript{\rm 2},
    Liu Liu\textsuperscript{\rm 3},
    Fan Huang\textsuperscript{\rm 4},\\
    Zhizhong Su\textsuperscript{\rm 3},
    Wei Sui\textsuperscript{\rm 2}\thanks{Project leader.},
    Xinggang Wang\textsuperscript{\rm 1}\thanks{Corresponding author (\href{mailto:xgwang@hust.edu.cn}{xgwang@hust.edu.cn}).}
}
\affiliations{
    \textsuperscript{\rm 1}Huazhong University of Science and Technology\\
    \textsuperscript{\rm 2}D-Robotics,
    \textsuperscript{\rm 3}Horizon Robotics,
    \textsuperscript{\rm 4}Xiamen University\\[0.2em]
    Code \& Models: \href{https://github.com/hustvl/FasterWAM}{\textcolor{niceblue}{hustvl/FasterWAM}}
}

\begin{document}

\maketitle

\begin{abstract}
World Action Models (WAMs) improve robot manipulation by learning how the environment evolves beyond the current observation. 
However, existing approaches face a fundamental dilemma: Joint-WAMs preserve future-aware representations during inference but incur prohibitive computation costs, while efficient alternatives remove future modeling at inference time and may lose the robustness benefits of temporal reasoning.
In this work, we revisit the role of future representations in WAMs and show that inference-time future conditioning is critical for generalization under distribution shifts. 
This observation motivates Faster-WAM, an efficient future-conditioning WAM that preserves future representations while avoiding expensive video–action interaction.
Faster-WAM introduces a sparse future-conditioning framework that computes future representations once and selectively reuses them throughout action denoising. 
Specifically, we propose SparseMoT to replace ubiquitous layer-wise fusion with selective video--action interaction at a compact subset of network stages, and Interval KV-Fusion to aggregate multi-depth future representations without increasing attention complexity.
Experiments demonstrate that Faster-WAM achieves a substantially better performance–efficiency trade-off than existing WAMs. 
On the out-of-distribution LIBERO-Plus benchmark, Faster-WAM improves success rate from 49.14\% to 73.57\% compared with Fast-WAM, while running 2.21× faster than Joint-WAM. 
It further achieves state-of-the-art performance on LIBERO and RoboTwin 2.0, while demonstrating strong robustness in real-world manipulation.
\end{abstract}

\section{Introduction}
\label{sec:intro}
General-purpose robot intelligence requires not only recognizing the current environment but also anticipating how the world will evolve after interaction.
While recent Vision-Language-Action (VLA) models~\cite{black2024pi_0,intelligence2025pi_,kim2024openvla} have achieved impressive progress in robotic manipulation, most existing approaches predict actions primarily from current visual observations and language instructions, without explicitly modeling future scene dynamics.
World Action Models (WAMs)~\cite{tian2025predictive,ye2026world} address this limitation by augmenting action prediction with future visual modeling.
By learning how objects move and scenes evolve under robot interaction, WAMs provide policies with temporal representations beyond the current observation.
However, an important question remains unresolved:
\textit{\textbf{Are future predictions merely useful as a training signal, or do future representations provide essential information during inference?}}

Existing WAMs provide two different answers.
Joint-WAMs~\cite{lingbot-va2026,bi2025motusunifiedlatentaction} couple future video generation and action prediction through shared denoising, allowing the action branch to access evolving future representations during inference.
However, repeatedly running the video branch and performing dense video--action interaction introduces substantial computational overhead.
Fast-WAM~\cite{yuan2026fastwam} explores the opposite direction by using future modeling only during training and removing future representations during inference.
Although this design significantly improves efficiency and achieves competitive in-distribution performance~\cite{liu2023libero,chen2025robotwin}, it raises a fundamental limitation: without inference-time future conditioning, the policy may lose temporal information required to handle unseen environments.

\begin{figure}[t]
\centering
\includegraphics[width=\columnwidth]{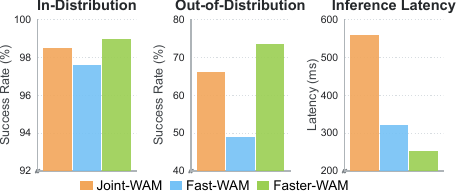}
\caption{
Inference-time future conditioning is critical for robust WAMs.
Compared with Joint-WAM (a controlled implementation representing Joint-WAMs), Fast-WAM improves efficiency by removing future conditioning at inference, yet suffers a marked performance drop under distribution shift.
Motivated by this finding, Faster-WAM efficiently preserves inference-time future conditioning, achieving strong OOD robustness at lower latency.
}
\label{fig:faster_wam_motivation}
\end{figure}

To investigate this question, we evaluate WAMs under distribution shift~\cite{fei25libero-plus} and observe that removing future representations at inference substantially harms robustness.
Specifically, as shown in Fig.~\ref{fig:faster_wam_motivation}, Fast-WAM exhibits a significant performance degradation in out-of-distribution (OOD) settings compared with Joint-WAM, suggesting that future representations are not merely an auxiliary training objective but an important source of generalizable temporal knowledge.
This finding leads to a new design principle for WAMs:
\textit{\textbf{Future representations should be preserved at inference, but their interaction with action prediction must become selective and efficient.}}

Based on this principle, we propose Faster-WAM, an efficient future-conditioning World Action Model that maintains inference-time future conditioning while redesigning video--action interaction.
Instead of repeatedly executing the video branch, Faster-WAM computes future representations once and reuses them through cached intermediate representations during action denoising.
To achieve efficient future conditioning, Faster-WAM introduces two complementary mechanisms.
First, SparseMoT reduces unnecessary computation by concentrating video--action interaction at a compact subset of stages while performing lightweight action-only refinement between successive interactions.
Second, Interval KV-Fusion aggregates future representations from multiple video depths within each interaction interval, providing richer temporal information without increasing attention complexity.

Extensive experiments demonstrate that Faster-WAM achieves a superior balance between robustness and efficiency.
On the OOD LIBERO-Plus benchmark~\cite{fei25libero-plus}, Faster-WAM achieves a 73.57\% success rate compared with 49.14\% for Fast-WAM, while achieving a $2.21\times$ inference speedup over Joint-WAM.
It also achieves state-of-the-art performance on LIBERO and RoboTwin 2.0, while demonstrating strong robustness in real-world manipulation.

Our contributions are summarized as follows:
\begin{itemize}
\item We identify inference-time future conditioning as an important factor for WAM generalization, showing that future representations provide robustness beyond their role as a training objective.
\item We propose Faster-WAM, a future-conditioning framework that preserves inference-time temporal representations through sparse and efficient video--action interaction.
\item We introduce SparseMoT and Interval KV-Fusion, enabling selective access to multi-level future representations without the computational cost of dense interaction.
\item Extensive experiments demonstrate state-of-the-art performance on in- and out-of-distribution benchmarks with improved inference efficiency.
\end{itemize}

\section{Related Work}
\label{sec:related_work}
\paragraph{Vision-Language-Action Models.}
VLAs have become a dominant framework for vision-language-conditioned robot manipulation~\cite{intelligence2025pi_,kim2024openvla,bjorck2025gr00t}.
Models such as RT-2~\cite{zitkovich2023rt} and $\pi_0$~\cite{black2024pi_0} show that pretrained vision-language backbones~\cite{driess2023palm,beyer2024paligemma} can transfer broad semantic knowledge to robot control.
Recent work further scales robot data~\cite{o2024open,bu2025agibot,khazatsky2024droid} and improves action interfaces through tokenization, diffusion, and flow matching~\cite{pertsch2025fast,liu2025rdt}.
Despite these advances, most VLAs predict actions without explicitly modeling future scene evolution, motivating policies that incorporate future visual dynamics~\cite{ye2026world}.

\paragraph{World Action Models.}
WAMs address this limitation by incorporating future visual dynamics into robot policy learning, building on increasingly capable video-generation priors~\cite{wan2025wan,seedance2026seedance,gao2026dreamdojo}.
Early predictive policies~\cite{du2023learning,tian2025predictive,hu2024video} treat imagined visual futures as intermediate planning objects, from which actions are subsequently recovered.
Joint-WAMs~\cite{lingbot-va2026,bi2025motusunifiedlatentaction,kim2026cosmos} instead couple future-video and action generation, allowing action prediction to access evolving future representations. 
However, such joint modeling remains expensive at inference~\cite{nextforcing}, as it requires iterative video computation and dense video--action interaction.
Fast-WAM~\cite{yuan2026fastwam} studies the controlled alternative of retaining future modeling only as a training objective and removing future representations at inference.
Light-WAM~\cite{li2026light} follows the same training-only route while further reducing deployment cost through a compact video backbone and direct action decoding.
However, whether future representations can be discarded entirely at inference without compromising robustness under distribution shift remains underexplored.

Concurrent Efficient-WAM~\cite{li2026efficient} also seeks to retain future latents efficiently, focusing on compressing future generation through a distilled video expert, low-resolution future tokens, and asymmetric denoising.
Faster-WAM instead takes the interaction between future representations and action generation as its primary object of design.
Built on a reusable future context derived from a single video-expert pass, it selectively exposes multi-depth future representations to the action pathway during action denoising.
Thus, rather than treating acceleration as the endpoint, Faster-WAM treats future representations as potentially valuable under distribution shift and pursues efficiency by controlling when and how they condition action generation.

\section{Method}
\label{sec:method}
\begin{figure*}[t]
    \centering
    \includegraphics[width=0.9\textwidth]{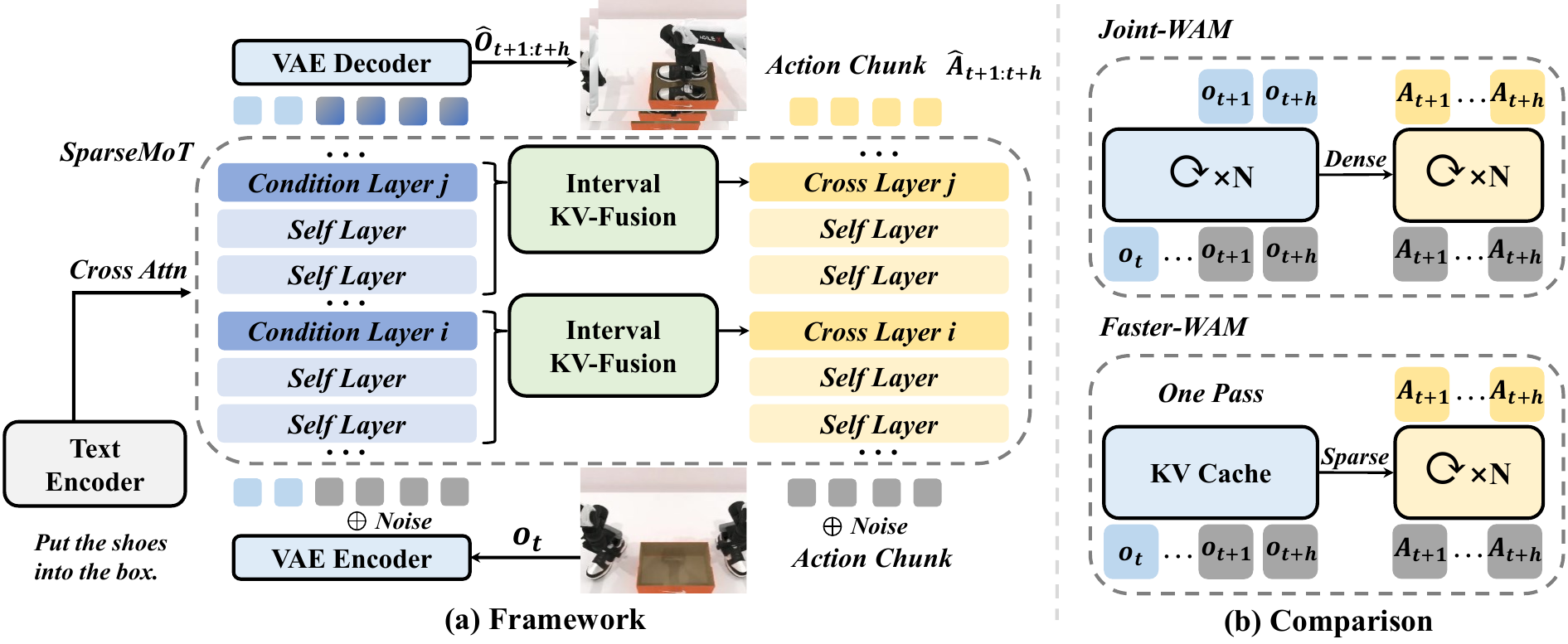}
    \caption{Overview of Faster-WAM. (a) Overall framework of Faster-WAM, featuring SparseMoT for selective video--action interaction and Interval KV-Fusion for aggregating multi-level future representations. (b) Inference comparison between Joint-WAM and Faster-WAM, contrasting iterative dense coupling with one-pass future-context caching and sparse interaction.}
    \label{fig:faster_wam_framework}
\end{figure*}

\subsection{Problem Formulation}
\label{sec:problem_formulation}

We consider language-conditioned visuomotor control from demonstrations.
At control step $t$, the policy observes an image $o_t$, a language instruction $l$, and a proprioceptive state $s_t$, and predicts an action chunk $A_t=a_{t+1:t+H}$ of horizon $H$.
A WAM parameterized by $\theta$ defines its action policy through an internal visual interface:
\begin{equation}
    \pi_\theta
    =
    p_\theta\!\left(
        A_t
        \mid s_t,l,\mathcal{R}_t^v
    \right),
    \label{eq:wam_policy}
\end{equation}
where $\mathcal{R}_t^v$ is the internal visual representation derived from the current observation $o_t$ by the video branch.
Let $\mathcal{E}_v$ denote the visual encoder, $z_t^0=\mathcal{E}_v(o_t)$ the current-observation latent, and $Z_t$ the future-video latents.
Joint-WAMs~\cite{ye2026world,bi2025motusunifiedlatentaction} jointly update future-video and action states, yielding an evolving video representation at step $k$:
\begin{equation}
    \mathcal{R}_{t,k}^{v,\mathrm{joint}}
    =
    G_v\!\left(z_t^0,Z_t^{(k)},l\right).
    \label{eq:joint_wam_representation}
\end{equation}
Here, $G_v$ denotes the video-side representation map.
As $Z_t^{(k)}$ evolves, the dense Joint-WAM used for comparison recomputes this representation and repeats cross-branch interaction at every step.
Fast-WAM~\cite{yuan2026fastwam} instead removes future slots at inference:
\begin{equation}
    \mathcal{R}_t^{v,\mathrm{fast}}
    =
    G_v\!\left(z_t^0,l\right).
    \label{eq:fast_wam_representation}
\end{equation}
The resulting interface can be reused throughout action generation, but is constructed without explicit future temporal slots.
Faster-WAM instead constructs a fixed future-aware interface $\overline{\mathcal{R}}_t^v$ in one video-expert pass and reuses it through sparse interaction during action generation (Fig.~\ref{fig:faster_wam_framework}(b)).

\subsection{Faster-WAM}
\label{sec:faster_wam_framework}

\paragraph{Overview.}
As illustrated in Fig.~\ref{fig:faster_wam_framework}(a), Faster-WAM couples a video expert initialized from a pretrained video generator~\cite{wan2025wan} with an action expert through a Mixture-of-Transformers (MoT) architecture~\cite{liang2024mixture}.
The architecture comprises $L$ aligned stages, each pairing the corresponding video and action layers.
Language and proprioception are supplied as shared conditioning signals to both the video and action experts.
A single video pass produces a layer-wise attention key/value (K/V) hierarchy.
Interval KV-Fusion turns it into compact action-facing summaries for SparseMoT to expose at selected stages, leaving the remaining stages for action-only refinement.

\paragraph{One-pass Future Conditioning.}
Prior work~\cite{pai2025mimicvideo,ma2026dit4dit} shows that control-relevant states can be extracted from high-noise video latents without completing denoising.
We formulate both branches with flow matching, taking $\tau=0$ as clean data and $\tau=1$ as Gaussian noise, and denote the interpolated future latent at video flow time $\tau_v$ by $Z_{t,\tau_v}$ (Eq.~\ref{eq:flow_interpolation}).
At the noisy endpoint, the video expert processes Gaussian future slots together with the clean current-frame anchor $z_t^0$ and language.
Estimating the flow-matching direction for these future-video latents requires reasoning about plausible scene dynamics, so the resulting hidden states can encode future-aware cues without reconstructing a rollout.
To expose these cues to the action expert, we retain from each video attention block at flow time $\tau_v$ the key/value projections $K_{t,\tau_v,j}^v$ and $V_{t,\tau_v,j}^v$, whose token layout and shape are shared across depth.
Collecting them gives
\begin{equation}
    \mathcal{C}_{t,\tau_v}^v
    =
    \left\{
    \left(
    K_{t,\tau_v,j}^v,
    V_{t,\tau_v,j}^v
    \right)
    \right\}_{j=1}^{L}.
    \label{eq:video_context}
\end{equation}
This raw K/V hierarchy is the source from which the action-facing interface is constructed.
To make it reusable across action flow steps, we use asymmetric attention within the video stream: future slots may attend to the clean anchor and one another, while the anchor cannot attend to them.
Across experts, action queries may read video features, whereas video queries cannot read action tokens.
Consequently, the video hierarchy is independent of the evolving action trajectory and can be constructed before action integration.
At inference, setting $Z_{t,1}$ to Gaussian noise $\epsilon_t^v$ yields the fixed raw hierarchy $\mathcal{C}_{t,1}^v$ in one video pass.

\paragraph{SparseMoT.}
In a conventional dense MoT, video and action features interact at every aligned stage.
Even with a precomputed video hierarchy, this cross-branch attention repeats across all $L$ stages at every action flow step.
SparseMoT reduces this repeated cost by restricting video access to the interaction set
\begin{equation}
    \begin{array}{c}
    \mathcal{J}
    =
    \{j_1,\ldots,j_M\}
    \subseteq
    \{1,\ldots,L\},\\
    1\le j_1<\cdots<j_M\le L.
    \end{array}
    \label{eq:interaction_layers}
\end{equation}
We select $\mathcal{J}$ at a fixed layer stride and reuse it at every action flow step; $M$ therefore counts the video-reading stages per action evaluation, with $M=L$ recovering dense~MoT.
For notational simplicity, we consider a fixed control step~$t$, video flow time~$\tau_v$, and action flow time~$\tau_a$, and omit these indices below.
Let $Q_j^a$, $K_j^a$, and $V_j^a$ denote the action-token query, key, and value projections at stage $j$.
At $j_m\in\mathcal{J}$, $(\widehat{K}_{j_m}^v,\widehat{V}_{j_m}^v)$ is the fused video pair summarizing its preceding depth interval (Eq.~\ref{eq:interval_kv_fusion}).
Its head dimensions match those of the action K/V projections, enabling the update
\begin{equation}
    \widetilde{X}_{j_m}^a
    =
    \mathrm{Attn}\!\left(
        Q_{j_m}^a,
        [\widehat{K}_{j_m}^v;K_{j_m}^a],
        [\widehat{V}_{j_m}^v;V_{j_m}^a]
    \right),
    \label{eq:sparse_cross_attention}
\end{equation}
where $\mathrm{Attn}$ is standard attention, $[\,;\,]$ concatenates tokens, and $\widetilde{X}_{j_m}^a$ is the action output combining the future summary with the current action state.
For $j\notin\mathcal{J}$, action-only self-attention and residual/feed-forward updates carry previously injected future information through the action state without reading video K/V again.
Thus all $L$ action stages remain active, and only cross-branch communication is sparse.

\paragraph{Interval KV-Fusion.}
SparseMoT reduces how often the action pathway reads video context, while the video expert continues to transform its representation at the intervening depths.
If interaction stage~$j_m$ consumed only its own K/V pair, intermediate representations would not be directly exposed to the action pathway.
Alternatively, concatenating them would lengthen the action-attention context.
To address this, we propose Interval KV-Fusion, which aggregates the video K/V pairs accumulated within each interaction interval.
Specifically, we assign each selected stage~$j_m$ exactly one preceding interval $\mathcal{I}_m=\{j_{m-1}+1,\ldots,j_m\}$.
For each assigned interval~$\mathcal{I}_m$, we introduce softmax-normalized fusion weights $W^{\mathrm{fuse}}_{m,j}$ to aggregate the video representations across its stages without lengthening action attention.
Because the stage-wise K/V pairs share a common token layout and dimensionality, the fused pair for $j_m$ is
\begin{equation}
    \left(
    \widehat{K}_{j_m}^v,
    \widehat{V}_{j_m}^v
    \right)
    =
    \sum_{j\in\mathcal{I}_m}
    W^{\mathrm{fuse}}_{m,j}
    \left(
    K_j^v,
    V_j^v
    \right).
    \label{eq:interval_kv_fusion}
\end{equation}
This weighted sum combines the K/V information from all stages in $\mathcal{I}_m$ into a single pair for $j_m$, while preserving key--value correspondence and the sequence length of one video stage.
Restoring $t$ and $\tau_v$, the $M$ fused pairs form the action-facing interface
\begin{equation}
    \widehat{\mathcal{C}}_{t,\tau_v}^v
    =
    \left\{
    \left(
    \widehat{K}_{t,\tau_v,j_m}^v,
    \widehat{V}_{t,\tau_v,j_m}^v
    \right)
    \right\}_{m=1}^{M}.
    \label{eq:fused_video_context}
\end{equation}
At inference, setting $\tau_v=1$ produces the fixed interface $\widehat{\mathcal{C}}_{t,1}^v$, which is supplied to the action expert as reusable future-aware context throughout action integration.
Overall, Interval KV-Fusion preserves multi-depth future context under SparseMoT without increasing attention complexity.

\paragraph{Joint Training.}
Faster-WAM jointly learns the video and action flow fields through flow matching~\cite{lipman2022flow}.
For each training example, the video and action flow times $\tau_v$ and $\tau_a$ are sampled independently, together with Gaussian noise samples $\epsilon_t^v$ and $\epsilon_t^a$:
\begin{equation}
    \begin{array}{rcl}
        Z_{t,\tau_v}
        &=&
        (1-\tau_v)Z_t+\tau_v\epsilon_t^v,\\
        A_{t,\tau_a}
        &=&
        (1-\tau_a)A_t+\tau_a\epsilon_t^a.
    \end{array}
    \label{eq:flow_interpolation}
\end{equation}
Given $(Z_{t,\tau_v},\tau_v,z_t^0,l)$, the video expert predicts the video flow $\widehat{u}_t^v$ for the future slots while producing the raw K/V hierarchy $\mathcal{C}_{t,\tau_v}^v$.
Interval KV-Fusion converts this hierarchy into $\widehat{\mathcal{C}}_{t,\tau_v}^v$, which conditions the action-flow predictor $F_a$ through SparseMoT at the selected stages:
\begin{equation}
    \widehat{u}_t^a
    =
    F_a\!\left(
        A_{t,\tau_a},
        \tau_a
        \mid
        s_t,l,\widehat{\mathcal{C}}_{t,\tau_v}^v
    \right).
    \label{eq:conditioned_action_flow}
\end{equation}
With targets $u_t^v=\epsilon_t^v-Z_t$ and $u_t^a=\epsilon_t^a-A_t$, the joint objective is
\begin{equation}
    \begin{array}{rcl}
    \mathcal{L}
    &=&
    \lambda_v
    \mathrm{E}
    \left[
        W_v^{\mathrm{flow}}(\tau_v)
        \left\|
        \widehat{u}_t^v-u_t^v
        \right\|_2^2
    \right]
    \\[2pt]
    &&+
    \lambda_a
    \mathrm{E}
    \left[
        W_a^{\mathrm{flow}}(\tau_a)
        \left\|
        \widehat{u}_t^a-u_t^a
        \right\|_2^2    
    \right].
    \end{array}
    \label{eq:joint_training_objective}
\end{equation}
The video and action flow-matching MSE losses are weighted separately by the flow-time-dependent factors $W_v^{\mathrm{flow}}(\tau_v)$ and $W_a^{\mathrm{flow}}(\tau_a)$, respectively, while $\lambda_v$ and $\lambda_a$ balance the overall contributions of the two branches.
The video term supervises future-video dynamics, whereas the action term encourages the fused hierarchy to retain control-relevant information.
Independently sampling $\tau_v$ and $\tau_a$ exposes the action pathway to diverse combinations of video and action noise levels.

\paragraph{Efficient Inference.}
At deployment, Faster-WAM initializes the future slots and action state from Gaussian noise.
It evaluates the video expert once at $\tau_v=1$ and applies Interval KV-Fusion to the resulting raw hierarchy, yielding the cached action-facing interface $\widehat{\mathcal{C}}_{t,1}^v$.
At each action-flow step, the action expert reuses this cache through SparseMoT, without updating or decoding the future-video latents.
As illustrated in Fig.~\ref{fig:faster_wam_framework}(b), this replaces $N$ dense joint video--action evaluations with one video-side pass followed by $N$ sparse, cache-conditioned action evaluations.

\begin{table*}[t]
\centering
\newcommand{\liberotablebody}{%
{\small
\setlength{\tabcolsep}{2pt}
\begin{tabular}{lcccccc}
\toprule
Method & P.T. & Spa. & Obj. & Goa. & Lon. & Avg. \\
\midrule
$\pi_{0.5}$~\cite{intelligence2025pi_} & \cmark & 98.8 & 98.2 & 98.0 & 92.4 & 96.9 \\
LingBot-VA~\cite{lingbot-va2026} & \cmark & 98.5 & 99.6 & 97.2 & 98.5 & \underline{98.5} \\
Motus~\cite{bi2025motusunifiedlatentaction} & \cmark & 96.8 & 99.8 & 96.6 & 97.6 & 97.7 \\
Fast-WAM~\cite{yuan2026fastwam} & \xmark & 98.2 & 100.0 & 97.0 & 95.2 & 97.6 \\
Joint-WAM & \xmark & 99.6 & 99.4 & 98.2 & 96.8 & \underline{98.5} \\
\rowcolor[HTML]{F2F2F2}
\textbf{Faster-WAM (Ours)} & \xmark & 99.6 & 99.8 & 98.2 & 98.2 & \textbf{99.0} \\
\bottomrule
\end{tabular}
}
}

\newcommand{\robotwintablebody}{%
{\small
\setlength{\tabcolsep}{4pt}
\begin{tabular}{lcccc}
\toprule
Method & P.T. & Clean & Rand. & Avg. \\
\midrule
$\pi_{0.5}$~\cite{intelligence2025pi_} & \cmark & 82.7 & 76.8 & 79.8 \\
Motus~\cite{bi2025motusunifiedlatentaction} & \cmark & 88.7 & 87.0 & 87.9 \\
LingBot-VA~\cite{lingbot-va2026} & \cmark & 92.9 & 91.5 & \underline{92.2} \\
Fast-WAM~\cite{yuan2026fastwam} & \xmark & 91.9 & 91.8 & 91.9 \\
Joint-WAM & \xmark & 90.8 & 90.3 & 90.6 \\
\rowcolor[HTML]{F2F2F2}
\textbf{Faster-WAM (Ours)} & \xmark & 92.8 & 92.3 & \textbf{92.6} \\
\bottomrule
\end{tabular}
}
}

\newcommand{\liberoplustablebody}{%
{\small
\setlength{\tabcolsep}{6pt}
\begin{tabular}{lccccccccc}
\toprule
Method & P.T. & Camera & Robot & Lang. & Light & Backg. & Noise & Layout & Avg. \\
\midrule
UniVLA~\cite{bu2025univla} & \cmark & 1.8 & 46.2 & 69.6 & 69.0 & 81.0 & 21.2 & 31.9 & 42.9 \\
OpenVLA-OFT~\cite{kim2025fine} & \cmark & 56.4 & 31.9 & 79.5 & 88.7 & 93.3 & 75.8 & 74.2 & \underline{69.6} \\
$\pi_0$~\cite{black2024pi_0} & \cmark & 13.8 & 6.0 & 58.8 & 85.0 & 81.4 & 79.0 & 68.9 & 53.6 \\
$\pi_0$-Fast~\cite{pertsch2025fast} & \cmark & 65.1 & 21.6 & 61.0 & 73.2 & 73.2 & 74.4 & 68.8 & 61.6 \\
WorldVLA~\cite{cen2025worldvla} & \cmark & 0.1 & 27.9 & 41.6 & 43.7 & 17.1 & 10.9 & 38.0 & 25.0 \\
Fast-WAM~\cite{yuan2026fastwam} & \xmark & 18.8 & 45.7 & 70.1 & 83.2 & 45.7 & 29.8 & 62.7 & 49.1 \\
Joint-WAM & \xmark & 37.5 & 64.5 & 93.0 & 95.0 & 55.9 & 47.3 & 79.6 & 66.3 \\
\rowcolor[HTML]{F2F2F2}
\textbf{Faster-WAM (Ours)} & \xmark & 53.8 & 71.6 & 94.7 & 96.3 & 61.3 & 63.6 & 79.1 & \textbf{73.6} \\
\bottomrule
\end{tabular}
}
}

\captionbox{Success rates (\%) on LIBERO. P.T. denotes embodied pretraining. The best and second-best average results are shown in bold and underlined, respectively.\label{tab:libero}}[0.48\textwidth][c]{\liberotablebody}
\hfill
\captionbox{Success rates (\%) on RoboTwin 2.0. P.T. denotes embodied pretraining. The best and second-best average results are shown in bold and underlined, respectively.\label{tab:robotwin}}[0.48\textwidth][c]{\robotwintablebody}
\par
\captionbox{Success rates (\%) on LIBERO-Plus across seven distribution shifts. P.T. denotes embodied pretraining. The best and second-best average results are shown in bold and underlined, respectively.\label{tab:libero_plus}}[\textwidth][c]{\liberoplustablebody}
\end{table*}

\section{Experiment}
\label{sec:experiment}
\subsection{Implementation Details}

Faster-WAM is built upon Wan2.2-5B~\cite{wan2025wan}, from which we initialize the video DiT, text encoder, and VAE.
Continuous actions are modeled by a 30-layer action Transformer with a hidden width of 1024.
The policy predicts 32 actions at each planning step, while the corresponding visual sequence contains nine frames sampled at a temporal stride of four.
Observations from different cameras are spatially assembled before VAE encoding.
In our notation, the 30 action layers define $L=30$ aligned stages; interacting every 4 stages yields $M=8$ selected stages, while the remainder perform lightweight action-only updates.
Interval KV-Fusion aggregates the video representations associated with each interaction interval.
Video and action prediction are trained with a common flow-matching objective.
We optimize all models using AdamW with a learning rate of $1\times10^{-4}$ and weight decay of 0.01, followed by cosine learning-rate decay.
Training is performed in BF16 with the maximum gradient norm set to 1.0.
At test time, actions are obtained using 10 integration steps and a guidance scale of 1.0.

For controlled comparison~\cite{yuan2026fastwam}, we implement Joint-WAM, Fast-WAM, and Faster-WAM with the same pretrained video backbone, tokenization, training data, flow-matching objective, optimization recipe, and action-sampling settings.
Joint-WAM jointly denoises future-video and action latents with dense interaction, Fast-WAM removes future temporal slots at inference, and Faster-WAM reuses a one-pass future context through sparse interaction.
This shared implementation isolates the effect of inference-time future conditioning and video--action interaction.

\subsection{Experiment Setup}

We evaluate Faster-WAM on LIBERO, RoboTwin 2.0, the OOD LIBERO-Plus benchmark, and real-world dual-arm manipulation tasks, using success rate as the primary metric.

\paragraph{LIBERO.}
LIBERO~\cite{liu2023libero} is a standard benchmark for language-conditioned manipulation, comprising four suites that cover spatial relations, object-centric skills, goal-conditioned tasks, and long-horizon behaviors.
Each suite contains 10 tasks and 500 expert demonstrations.
We train for 10 epochs with a global batch size of 128 and evaluate each task over 50 rollouts.

\paragraph{RoboTwin 2.0.}
RoboTwin 2.0~\cite{chen2025robotwin} is a large-scale bimanual manipulation benchmark with more than 50 tasks requiring coordinated dual-arm control under diverse scene conditions.
We train a single policy on 2,500 clean and 25,000 randomized demonstrations for 5 epochs with a global batch size of 1,024.
Each task is evaluated over 100 trials in both clean and randomized settings.

\paragraph{LIBERO-Plus.}
LIBERO-Plus~\cite{fei25libero-plus} extends the original LIBERO tasks to evaluate robustness under conditions not observed during standard training.
It introduces seven types of variation involving camera viewpoints, robot initial states, language instructions, lighting, backgrounds, sensor noise, and object layouts.
We directly evaluate the LIBERO-trained policies without additional training.

\paragraph{Real-World Evaluation.}
We conduct real-world experiments on a dual-arm robot platform equipped with two Piper 6-DoF manipulators.
We consider four tasks: \textit{Pick Strawberries}, \textit{Build Tower}, \textit{Store Boxes}, and \textit{Stack Plates}.
Together, these tasks require fine-grained grasping, precise spatial alignment, dual-arm coordination, and multi-object manipulation, providing a diverse evaluation of real-world policy performance.
We collect 400 demonstrations per task and jointly train a single policy for 5 epochs with a global batch size of 512.
Each task is evaluated over 30 trials under the standard setting.
To assess OOD robustness, we further evaluate \textit{Pick Strawberries} under three conditions absent from the training demonstrations: altered lighting, novel backgrounds, and unseen distractor objects.

\begin{figure}[t]
\centering
\includegraphics[width=\columnwidth]{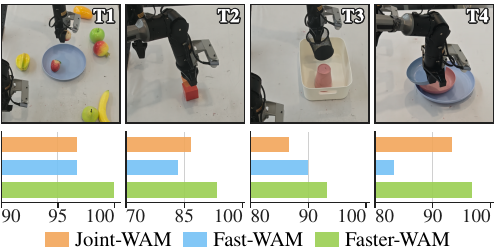}
\caption{Real-world success rates on four tasks: T1, Pick Strawberries; T2, Build Tower; T3, Store Boxes; and T4, Stack Plates. Bars from top to bottom show the success rates (\%) of Joint-WAM, Fast-WAM, and Faster-WAM, respectively.}
\label{fig:real_world_standard}
\end{figure}

\subsection{Main Results}

\paragraph{LIBERO.}
Table~\ref{tab:libero} summarizes the results across the four LIBERO suites.
Without embodied pretraining, Faster-WAM achieves an average success rate of 99.0\% and maintains at least 98.2\% on every suite.
Notably, all three WAM variants achieve average success rates above 97\%, reflecting their strong performance on standard LIBERO.
Despite this highly competitive regime, Faster-WAM achieves the strongest overall performance, outperforming both Joint-WAM and Fast-WAM in our controlled comparison and establishing its effectiveness under in-distribution evaluation.
The subsequent distribution-shift evaluations provide a more discriminative test of whether retaining future representations improves policy robustness.

\paragraph{RoboTwin 2.0.}
Table~\ref{tab:robotwin} demonstrates the performance of the compared methods under the clean and randomized evaluation settings.
Faster-WAM achieves success rates of 92.8\% and 92.3\%, respectively, yielding the best average success rate of 92.6\% despite using no embodied pretraining.
It outperforms the pretrained LingBot-VA as well as both controlled WAM baselines, demonstrating strong performance on large-scale bimanual manipulation across both settings.
Together with the LIBERO results, these findings establish the strong and consistent in-distribution performance of Faster-WAM across diverse manipulation benchmarks.

\paragraph{LIBERO-Plus.}
Table~\ref{tab:libero_plus} presents the results across seven unseen distribution shifts, providing a more discriminative evaluation of the three WAM variants.
Although Fast-WAM achieves an average success rate of 97.6\% on standard LIBERO, its performance drops to 49.1\% on LIBERO-Plus.
Joint-WAM retains a higher average success rate of 66.3\%, consistent with the benefit of preserving future representations at inference.
Faster-WAM further raises the average success rate to 73.6\%, compared with 49.1\% for Fast-WAM, and achieves the best overall performance among all evaluated methods.
It surpasses Fast-WAM across all seven distribution shifts, while outperforming Joint-WAM on six shifts and achieving comparable performance on Layout.
These consistent gains across diverse perturbations demonstrate the effectiveness of Faster-WAM for robust OOD manipulation.
\paragraph{Real-World Evaluation.}
\begin{figure}[t]
\centering
\includegraphics[width=\columnwidth]{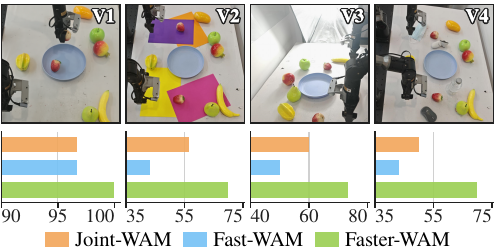}
\caption{Real-world OOD settings for Pick Strawberries under four conditions: V1, standard; V2, novel backgrounds; V3, altered lighting; and V4, unseen distractor objects. Bars from top to bottom show the success rates (\%) of Joint-WAM, Fast-WAM, and Faster-WAM, respectively.}
\label{fig:real_world_ood}
\end{figure}

Figures~\ref{fig:real_world_standard} and~\ref{fig:real_world_ood} present the standard and OOD results, respectively.
Under the standard setting, Faster-WAM outperforms both Joint-WAM and Fast-WAM on all four tasks, recording the highest overall success rate of 95.8\%, compared with 90.8\% for Joint-WAM and 88.3\% for Fast-WAM.
The advantage of Faster-WAM becomes more pronounced under the real-world distribution shifts shown in Fig.~\ref{fig:real_world_ood}.
Although Fast-WAM achieves a success rate of 96.7\% in the standard setting, its average performance drops to 45.6\% across the three unseen conditions, while Joint-WAM retains 55.6\%.
Faster-WAM achieves an OOD average success rate of 71.1\% and the best performance under every unseen condition.
Together with the LIBERO-Plus results, these findings confirm that Faster-WAM's robustness extends to real-world manipulation.
\begin{table}[t]
\centering
{\small
\setlength{\tabcolsep}{2pt}
\begin{tabular}{l*{4}{>{\centering\arraybackslash}p{32pt}}}
\toprule
Model & \makebox[32pt][c]{VAE Enc.} & Visual & Action & Overall \\
\midrule
Joint-WAM & 10.47 & -- & -- & 559.84 \\
Fast-WAM & 10.47 & 27.67 & 276.56 & 320.97 \\
Faster-WAM & 10.55 & 43.04 & 192.11 & \textbf{252.95} \\
\bottomrule
\end{tabular}
}
\caption{Overall and component-wise inference latency measurements (ms), averaged over 10 runs after 5 warm-up iterations. Dashes indicate that visual and action latency cannot be separated under joint video--action denoising.}
\label{tab:inference_latency}
\end{table}

\paragraph{Inference Latency.}
Table~\ref{tab:inference_latency} compares the inference latency of the three WAM variants measured on an NVIDIA L20 GPU.
All measurements use LIBERO-like image inputs at a resolution of $224\times448$ and 10 denoising steps.
Joint-WAM requires 559.84\,ms because it repeatedly updates the video and action branches through joint denoising.
By constructing the future-aware visual context once and reusing it through sparse video--action interaction throughout action denoising, Faster-WAM reduces the overall latency to 252.95\,ms, yielding a $2.21\times$ speedup over Joint-WAM.
Notably, despite retaining future representations at inference time, Faster-WAM is also faster than Fast-WAM, which requires 320.97\,ms.
The latency breakdown explains this advantage: although constructing the richer visual context increases the one-time visual latency from 27.67\,ms to 43.04\,ms, SparseMoT reduces the repeatedly incurred action-denoising latency from 276.56\,ms to 192.11\,ms, more than compensating for this overhead.
Together with the preceding OOD results, these findings establish the superior performance--efficiency trade-off of Faster-WAM.
\newcommand{\futureconditionrows}{%
Faster-WAM & 53.75 & 71.57 & 94.67 & 96.27 & 61.29 & 63.57 & 79.09 & \textbf{73.57} \\
C-O Faster-WAM & 16.25 & 50.25 & 77.05 & 85.09 & 44.09 & 32.95 & 62.73 & 51.00 \\
Fast-WAM & 18.75 & 45.69 & 70.08 & 83.23 & 45.70 & 29.84 & 62.73 & 49.14 \\
}

\newcommand{\componentablationrows}{%
Faster-WAM & 53.75 & 71.57 & 94.67 & 96.27 & 61.29 & 63.57 & 79.09 & \textbf{73.57} \\
w/o KV-Fusion & 47.50 & 66.50 & 92.21 & 98.14 & 54.84 & 54.26 & 83.64 & 69.99 \\
w/o SparseMoT & 48.10 & 77.05 & 95.95 & 98.41 & 58.73 & 47.87 & 73.91 & 69.78 \\
Joint-WAM & 37.50 & 64.47 & 93.03 & 95.03 & 55.91 & 47.29 & 79.55 & 66.27 \\
}

\newcommand{\interactionsparsityrows}{%
Dense & 48.10 & 77.05 & 95.95 & 98.41 & 58.73 & 47.87 & 73.91 & 69.78 \\
Stride-2 & 43.75 & 73.60 & 93.44 & 96.27 & 62.90 & 56.59 & 83.18 & 71.65 \\
Stride-4 & 53.75 & 71.57 & 94.67 & 96.27 & 61.29 & 63.57 & 79.09 & \textbf{73.57} \\
Stride-7 & 48.33 & 69.54 & 91.39 & 95.03 & 60.22 & 56.20 & 83.64 & 71.05 \\
Stride-14 & 45.00 & 69.54 & 89.34 & 96.89 & 54.30 & 58.53 & 83.64 & 70.05 \\
}

\begin{table*}[t]
\centering
\captionbox{Success rates (\%) on LIBERO-Plus across seven distribution shifts. (a) Future conditioning ablation comparing Faster-WAM with its current-only (C-O) counterpart and Fast-WAM. (b) Cumulative component ablation progressively removing Interval KV-Fusion, SparseMoT, and one-pass reusable future context. (c) Interaction-stride ablation comparing dense interaction with increasingly sparse video--action interaction.\label{tab:ablation}\label{tab:future_condition}\label{tab:component_ablation}\label{tab:interaction_sparsity}}[\textwidth][c]{%
{\small
\setlength{\tabcolsep}{7pt}
\begin{tabular}{lcccccccc}
\toprule
\rowcolor[HTML]{F2F2F2}
\multicolumn{9}{c}{\textbf{(a) Future Conditioning Ablation}} \\
\midrule
Method & Camera & Robot & Lang. & Light & Backg. & Noise & Layout & Avg. \\
\midrule
\futureconditionrows
\midrule
\rowcolor[HTML]{F2F2F2}
\multicolumn{9}{c}{\textbf{(b) Cumulative Component Ablation}} \\
\midrule
Model & Camera & Robot & Lang. & Light & Backg. & Noise & Layout & Avg. \\
\midrule
\componentablationrows
\midrule
\rowcolor[HTML]{F2F2F2}
\multicolumn{9}{c}{\textbf{(c) Interaction Stride Ablation}} \\
\midrule
Interaction & Camera & Robot & Lang. & Light & Backg. & Noise & Layout & Avg. \\
\midrule
\interactionsparsityrows
\bottomrule
\end{tabular}
}
}
\end{table*}

\subsection{Ablation Study}
\paragraph{Future Conditioning.}
Table~\ref{tab:ablation}(a) isolates the contribution of future conditioning from that of the proposed interaction architecture.
We construct C-O Faster-WAM, a current-only counterpart that retains SparseMoT and Interval KV-Fusion but forms its reusable visual context without explicit future temporal slots.
C-O Faster-WAM is trained from scratch under the same protocol as Faster-WAM and achieves an average success rate of 51.00\%, compared with 49.14\% for Fast-WAM.
Restoring future temporal slots raises the average success rate from 51.00\% to 73.57\%, with improvements across all seven distribution shifts.
The improvements are particularly pronounced under camera variation and visual noise, where the success rates increase from 16.25\% to 53.75\% and from 32.95\% to 63.57\%, respectively.
Because Faster-WAM and its current-only counterpart share the same interaction design and training protocol, this consistent gap isolates the contribution of preserving future-aware representations at inference.
These results support our central claim that future modeling is not merely an auxiliary training signal: explicit access to future-aware context is critical for robust OOD manipulation.

\paragraph{Component Ablation.}
Table~\ref{tab:ablation}(b) cumulatively removes Interval KV-Fusion, SparseMoT, and one-pass reusable future context from Faster-WAM.
Removing Interval KV-Fusion reduces the average success rate from 73.57\% to 69.99\%, confirming that intermediate-depth representations provide complementary future information.
Further replacing SparseMoT with dense MoT yields 69.78\%, and subsequently reverting the one-pass reusable context to iterative joint denoising recovers Joint-WAM at 66.27\%.
Overall, these results support the complementary roles of multi-depth KV fusion, sparse video--action interaction, and reusable future context in robust OOD manipulation.

\paragraph{Interaction Sparsity.}
Table~\ref{tab:ablation}(c) studies how video--action interaction density affects performance.
Dense interaction uses a stride of 1, whereas Stride-$k$ performs video--action interaction every $k$ stages with action-only refinement in between.
Increasing the stride from 1 to 2 and 4 raises the average success rate from 69.78\% to 71.65\% and 73.57\%, respectively.
Further increasing the stride to 7 and 14 reduces the average success rate to 71.05\% and 70.05\%.
This rise-then-fall pattern reveals that moderately sparse interaction is preferable to both dense and overly sparse coupling: future context need not be introduced at every stage, but sufficient interaction opportunities remain necessary to guide action prediction.

\section{Conclusion}
\label{sec:conclusion}
In this paper, we reveal a key design principle for WAMs: future representations should be treated not merely as an auxiliary training signal, but as essential inference-time context for robust action prediction under distribution shift.
Guided by this principle, we introduce Faster-WAM, an efficient future-conditioning framework that preserves inference-time future representations while reducing redundant video--action interaction.
SparseMoT concentrates cross-branch interaction at selected stages, while Interval KV-Fusion aggregates multi-depth future information into a compact action-facing context.
Faster-WAM achieves state-of-the-art in-distribution performance on standard benchmarks and robust OOD generalization in both simulated and real-world manipulation, while substantially reducing inference latency.
For future work, designing learning-based strategies to determine the video-action interaction stages may be a promising direction.
Moreover, graph compilation and custom CUDA kernels may further provide complementary system-level acceleration.
We believe this principle can inspire the development of future WAMs that retain and effectively exploit future context for robust robot manipulation.

\bibliography{aaai2027}

\clearpage 
\appendix
\section{Appendix}
\label{sec:appendix}
\setcounter{table}{0}
\renewcommand{\thetable}{A\arabic{table}}
\subsection{Training Details of Faster-WAM}

The main paper presents the Faster-WAM architecture and its flow-matching objective, while leaving detailed training configurations and implementation choices to the appendix due to space constraints.
Here, we describe the common optimization recipe and dataset-specific data construction used for LIBERO, RoboTwin 2.0, and the real-world experiments.
Faster-WAM jointly learns future visual prediction and action generation by optimizing the video and action experts together with Interval KV-Fusion and the proprioceptive encoder, while task instructions are supplied through precomputed language embeddings.
Joint-WAM and Fast-WAM follow the same training recipe to ensure controlled comparisons.

Across all datasets, we train in BF16 with DeepSpeed ZeRO-1 and use AdamW with betas $(0.9,0.95)$, a learning rate of $1\times10^{-4}$, and a weight decay of $1\times10^{-2}$.
Following a 5\% linear warm-up, the learning rate decays to $1\times10^{-6}$ under a cosine schedule, and the maximum gradient norm is clipped to 1.0.
The video and action flow times are sampled independently from separate 1,000-timestep schedulers with a shift of 5.0.
Both branches use timestep-weighted mean-squared-error losses, with padded positions masked out and equal weights assigned to the two objectives.
All runs use a training seed of 42.
The shared hyperparameters are summarized in Table~\ref{tab:training_hyperparams}.

Each training sample contains 33 consecutive observation-state pairs together with 32 action steps.
Sampling the visual sequence at indices $[0,4,\ldots,32]$ yields one current and eight future multi-view observations paired with a 32-step action chunk.
For LIBERO, the two camera views are concatenated horizontally into a $224\times448$ input, and training runs for 10 epochs on 8 NVIDIA A800 GPUs with a global batch size of 128.
For RoboTwin and the real-world data, the main view is stacked above two horizontally concatenated wrist views to produce a $384\times320$ input; both are trained for 5 epochs on 32 NVIDIA A800 GPUs, with global batch sizes of 1,024 and 512, respectively.
The dataset-specific hardware, batch, and training configurations are summarized in Table~\ref{tab:dataset_training_config}.

\subsection{Effect of Video Denoising Steps}

The main paper demonstrates that retaining future temporal slots at inference is critical for robustness under distribution shift.
Faster-WAM constructs its reusable future context from a single video-expert pass at the noisy endpoint $\tau_v=1$, where the future slots are still initialized from Gaussian noise.
To examine whether applying additional video denoising steps to these future slots produces a more informative future context, we conduct an additional ablation using the same trained checkpoint.
The detailed results are summarized in Table~\ref{tab:video_steps_ablation}.

At inference, we vary the number of video denoising steps used to construct the reusable future-aware context, ranging from the default one-pass setting to ten video steps.
After the selected number of steps, the resulting context is fixed and reused throughout the same 10-step action denoising process, with all other inference settings unchanged.
As shown in Table~\ref{tab:video_steps_ablation}, the default one-pass setting achieves the highest average success rate of 73.57\%.
Using two video steps does not improve performance, yielding 73.24\%, while performance generally declines with further denoising and reaches 68.33\% at ten steps.
These results support our design rationale: estimating the flow direction at the noisy endpoint already requires reasoning about plausible scene dynamics, allowing the resulting representations to encode control-relevant future cues before the future latents are explicitly resolved.
Together with the current-only ablation in the main paper, this finding shows that future-aware context is important for robust action prediction, while iterative future reconstruction is unnecessary for effective future conditioning.

\subsection{Details of the Latency Comparison}
\label{sec:latency_details}

The main paper reports the inference latency of the three WAM variants. Due to space constraints, we provide further measurement details and run-to-run variability in the appendix, along with results for the w/o KV-Fusion and w/o SparseMoT configurations from the cumulative component ablation. Table~\ref{tab:latency_details} summarizes the detailed results.
\begin{table}[t]
\centering
\begin{tabular}{ll}
\toprule
Parameter & Value \\
\midrule
Distributed strategy & DeepSpeed ZeRO-1 \\
Precision & BF16 \\
Optimizer & AdamW, betas $(0.9, 0.95)$ \\
Learning rate & $1\times10^{-4}$ \\
Weight decay & $1\times10^{-2}$ \\
LR scheduler & 5\% warmup, cosine to $1\times10^{-6}$ \\
Gradient clipping & 1.0 \\
Flow schedulers & 1,000 timesteps, shift 5.0 \\
Loss weights & $\lambda_v=\lambda_a=1.0$ \\
Seed & 42 \\
\bottomrule
\end{tabular}
\caption{Common training hyperparameters.}
\label{tab:training_hyperparams}
\end{table}

\begin{table}[t]
\centering
\setlength{\tabcolsep}{2.5pt}
\begin{tabular}{lccc}
\toprule
Setting & LIBERO & RoboTwin & Real \\
\midrule
GPUs & 8 & 32 & 32 \\
Input views & 2 & 3 & 3 \\
Resolution ($H\times W$) & $224\times448$ & $384\times320$ & $384\times320$ \\
Visual sequence & $1+8$ & $1+8$ & $1+8$ \\
Visual stride & 4 & 4 & 4 \\
Action chunk & 32 & 32 & 32 \\
Epochs & 10 & 5 & 5 \\
Micro-batch/GPU & 16 & 8 & 16 \\
Grad. accum. & 1 & 4 & 1 \\
Global batch & 128 & 1,024 & 512 \\
\bottomrule
\end{tabular}
\caption{Dataset-specific training configurations. The visual sequence comprises one current and eight future observations.}
\label{tab:dataset_training_config}
\end{table}

\begin{table*}[t]
\centering
\captionbox{LIBERO-Plus success rates (\%) with different numbers of video denoising steps. All settings use the same Faster-WAM checkpoint and 10 action-integration steps, caching only the K/V hierarchy produced at the final video step. Avg. is computed over all tasks rather than over categories, with the best value shown in bold.\label{tab:video_steps_ablation}}[\textwidth][c]{%
\setlength{\tabcolsep}{8pt}
\begin{tabular}{lcccccccc}
\toprule
Video Steps & Camera & Robot & Lang. & Light & Backg. & Noise & Layout & Avg. \\
\midrule
1  & 53.75 & 71.57 & 94.67 & 96.27 & 61.29 & 63.57 & 79.09 & \textbf{73.57} \\
2  & 56.25 & 70.05 & 93.03 & 96.27 & 56.45 & 62.79 & 82.27 & 73.24 \\
3  & 50.42 & 69.54 & 95.08 & 95.65 & 58.06 & 60.85 & 82.27 & 72.38 \\
4  & 50.42 & 66.50 & 95.90 & 95.65 & 58.06 & 60.47 & 82.27 & 72.05 \\
5  & 43.75 & 68.53 & 94.67 & 94.41 & 57.53 & 58.91 & 83.18 & 70.72 \\
6  & 43.75 & 68.02 & 93.85 & 98.14 & 57.53 & 55.43 & 82.27 & 70.19 \\
7  & 48.33 & 68.53 & 91.80 & 94.41 & 56.99 & 55.43 & 83.64 & 70.39 \\
8  & 42.92 & 65.99 & 92.62 & 95.65 & 54.84 & 57.36 & 80.45 & 69.06 \\
9  & 41.67 & 63.96 & 91.80 & 94.41 & 52.69 & 56.59 & 80.00 & 67.86 \\
10 & 41.67 & 63.45 & 91.80 & 95.03 & 55.38 & 56.59 & 80.91 & 68.33 \\
\bottomrule
\end{tabular}
}
\end{table*}

\begin{table*}[t]
\centering
\begin{tabular}{lcccc}
\toprule
Model & VAE Enc. & Visual & Action & Overall \\
\midrule
Joint-WAM & $10.47 \pm 0.02$ & -- & -- & $559.84 \pm 7.61$ \\
Fast-WAM & $10.47 \pm 0.04$ & $27.67 \pm 0.44$ & $276.56 \pm 2.77$ & $320.97 \pm 2.86$ \\
Faster-WAM & $10.55 \pm 0.06$ & $43.04 \pm 0.04$ & $192.11 \pm 2.36$ & $252.95 \pm 2.42$ \\
w/o KV-Fusion & $10.48 \pm 0.02$ & $42.55 \pm 0.06$ & $191.61 \pm 2.10$ & $252.00 \pm 2.38$ \\
w/o SparseMoT & $10.51 \pm 0.03$ & $42.81 \pm 0.02$ & $278.08 \pm 2.93$ & $339.00 \pm 3.66$ \\
\bottomrule
\end{tabular}
\caption{Detailed inference latency (ms), reported as the mean $\pm$ standard deviation over 10 runs after 5 warm-up iterations. Dashes indicate that visual and action latency cannot be separated under joint video--action denoising.}
\label{tab:latency_details}
\end{table*}

All configurations use BF16 on a single NVIDIA L20 GPU with $224\times448$ LIBERO-like inputs, an action horizon of 32, and 10 action-denoising steps. We perform 5 warm-up iterations followed by 10 GPU-synchronized measurements and report the mean and standard deviation. The model and input seeds are fixed to 42 and 0, respectively. Image tensors, language embeddings, and proprioception are prepared before timing, while image preprocessing, language encoding, action denormalization, and gripper post-processing are excluded. VAE encoding measures the time required to encode the current observation into the visual latent used by the model. Visual latency includes video-side context preparation, the video-expert forward pass, and reusable K/V cache construction. Action latency includes action-latent initialization, all 10 denoising steps, and the final action transfer to the CPU. Overall latency covers the full inference entry point, including shared preparation overhead. Joint-WAM updates the video and action branches jointly, so only its overall latency is reported.
In the cumulative ablation, removing Interval KV-Fusion changes overall latency by less than 1\,ms, while further removing SparseMoT increases it from 252.00\,ms to 339.00\,ms. These results show that SparseMoT provides the primary efficiency gain, while Interval KV-Fusion introduces negligible measured overhead.

\subsection{Details of the Real-World Dataset}
\label{sec:real_world_dataset_details}

The main paper reports our evaluation of Faster-WAM on four real-world dual-arm manipulation tasks and provides an overview of the corresponding dataset.
Here, we provide further details on the task definitions, dataset composition, and observation format.
Following the naming and ordering used in the main paper, \textit{Pick Strawberries} (T1) requires the robot to identify the strawberries among other fruits and place them on the plate.
\textit{Build Tower} (T2) requires the robot to first move the red and orange blocks to the center of the table and then stack the orange block on top of the red block.
\textit{Store Boxes} (T3) requires the robot to clear the tabletop by placing the red cup and black pen holder into an open storage bin, beginning with whichever object is initially located on the right.
Finally, \textit{Stack Plates} (T4) requires the robot to first place the small red plate on the large plate and then stack the small blue plate on top of the red plate.

The combined dataset contains 1,600 demonstrations, evenly divided into 400 demonstrations per task, and 488,393 synchronized time steps recorded at 30 FPS, totaling approximately 4.52 hours.
Each demonstration records synchronized $480\times640$ RGB streams from the head, left-wrist, and right-wrist cameras.
Both proprioception and action are 14-dimensional, comprising six joint channels and one gripper channel per arm.
During training, the three views are resized and spatially assembled into the $384\times320$ composite observation described earlier in this appendix.
The OOD evaluation of \textit{Pick Strawberries} uses novel backgrounds, altered lighting, and unseen distractor objects, none of which appear in the training demonstrations.

\subsection{Detailed Results on RoboTwin 2.0}
\label{sec:robotwin_detailed_results}

The main paper reports average success rates on RoboTwin 2.0 under both the clean and randomized evaluation settings.
Here, we complement this summary by providing the corresponding per-task success rates for Faster-WAM and all compared methods.
The detailed results are presented in \mbox{Table~\ref{tab:robotwin_detailed}}.
\begin{table*}[t]
\centering
\captionbox{Per-task success rates (\%) on RoboTwin 2.0 under the clean and randomized evaluation settings. The best per-task and average results for each setting are shown in bold.\label{tab:robotwin_detailed}}[\textwidth][c]{%
\setlength{\tabcolsep}{1mm}
\begin{tabular}{@{}lcccccccccccc@{}}
\toprule
Task & \multicolumn{2}{c}{Faster-WAM} & \multicolumn{2}{c}{Fast-WAM} & \multicolumn{2}{c}{Joint-WAM} & \multicolumn{2}{c}{LingBot-VA} & \multicolumn{2}{c}{$\pi_{0.5}$} & \multicolumn{2}{c}{Motus} \\
\cmidrule(lr){2-3}\cmidrule(lr){4-5}\cmidrule(lr){6-7}\cmidrule(lr){8-9}\cmidrule(lr){10-11}\cmidrule(lr){12-13}
 & Clean & Rand. & Clean & Rand. & Clean & Rand. & Clean & Rand. & Clean & Rand. & Clean & Rand. \\
\midrule
Adjust Bottle & \textbf{100} & 99 & \textbf{100} & \textbf{100} & 98 & 99 & 90 & 94 & \textbf{100} & 99 & 89 & 93 \\
Beat Block Hammer & 97 & \textbf{98} & 99 & 97 & \textbf{100} & \textbf{98} & 96 & \textbf{98} & 96 & 93 & 95 & 88 \\
Blocks Ranking RGB & \textbf{100} & \textbf{100} & \textbf{100} & \textbf{100} & \textbf{100} & \textbf{100} & 99 & 98 & 92 & 85 & 99 & 97 \\
Blocks Ranking Size & 74 & 93 & \textbf{94} & \textbf{98} & 83 & 91 & \textbf{94} & 96 & 49 & 26 & 75 & 63 \\
Click Alarmclock & 99 & \textbf{100} & \textbf{100} & \textbf{100} & \textbf{100} & \textbf{100} & 99 & \textbf{100} & 98 & 89 & \textbf{100} & \textbf{100} \\
Click Bell & \textbf{100} & \textbf{100} & \textbf{100} & \textbf{100} & \textbf{100} & 98 & \textbf{100} & \textbf{100} & 99 & 66 & \textbf{100} & \textbf{100} \\
Dump Bin Bigbin & \textbf{97} & \textbf{98} & \textbf{97} & 96 & 95 & 95 & 89 & 96 & 92 & 97 & 95 & 91 \\
Grab Roller & \textbf{100} & \textbf{100} & \textbf{100} & \textbf{100} & \textbf{100} & \textbf{100} & \textbf{100} & \textbf{100} & \textbf{100} & \textbf{100} & \textbf{100} & \textbf{100} \\
Handover Block & 93 & 88 & 95 & 81 & 93 & \textbf{91} & \textbf{99} & 78 & 66 & 57 & 86 & 73 \\
Handover Mic & \textbf{100} & 99 & 99 & \textbf{100} & \textbf{100} & \textbf{100} & 94 & 96 & 98 & 97 & 78 & 63 \\
Hanging Mug & 61 & 44 & 58 & \textbf{62} & \textbf{71} & 56 & 40 & 28 & 18 & 17 & 38 & 38 \\
Lift Pot & \textbf{100} & \textbf{100} & \textbf{100} & \textbf{100} & \textbf{100} & \textbf{100} & \textbf{100} & 99 & 96 & 85 & 96 & 99 \\
Move Can Pot & 96 & 98 & 90 & 88 & \textbf{97} & \textbf{99} & 94 & 97 & 51 & 55 & 34 & 74 \\
Move Pillbottle Pad & 98 & 99 & \textbf{100} & 99 & 99 & \textbf{100} & 99 & 99 & 84 & 61 & 93 & 96 \\
Move Playingcard Away & \textbf{100} & \textbf{100} & \textbf{100} & \textbf{100} & \textbf{100} & \textbf{100} & \textbf{100} & 99 & 96 & 84 & \textbf{100} & 96 \\
Move Stapler Pad & 82 & 74 & 77 & 64 & 85 & 81 & \textbf{91} & 79 & 56 & 42 & 83 & \textbf{85} \\
Open Laptop & 95 & 99 & \textbf{98} & \textbf{100} & 89 & 92 & 92 & 94 & 90 & 96 & 95 & 91 \\
Open Microwave & 77 & 77 & 62 & 45 & 3 & 14 & 82 & 86 & 34 & 77 & \textbf{95} & \textbf{91} \\
Pick Diverse Bottles & \textbf{90} & 89 & 80 & 85 & 86 & 87 & 89 & 82 & 81 & 71 & \textbf{90} & \textbf{91} \\
Pick Dual Bottles & 98 & 98 & \textbf{100} & 96 & 98 & \textbf{99} & \textbf{100} & \textbf{99} & 93 & 63 & 96 & 90 \\
Place A2B Left & 96 & 94 & 95 & 93 & 96 & \textbf{96} & \textbf{97} & 93 & 87 & 82 & 88 & 79 \\
Place A2B Right & 94 & 93 & 93 & \textbf{99} & 95 & 95 & \textbf{97} & 95 & 87 & 84 & 91 & 87 \\
Place Bread Basket & 89 & \textbf{96} & 91 & 93 & 89 & 94 & \textbf{97} & 95 & 77 & 64 & 91 & 94 \\
Place Bread Skillet & 93 & 87 & 90 & \textbf{93} & 90 & \textbf{93} & \textbf{95} & 90 & 85 & 66 & 86 & 83 \\
Place Burger Fries & 98 & 97 & 96 & 99 & \textbf{100} & \textbf{100} & 97 & 95 & 94 & 87 & 98 & 98 \\
Place Can Basket & 76 & 73 & 71 & 69 & 50 & 23 & \textbf{81} & \textbf{84} & 62 & 62 & \textbf{81} & 76 \\
Place Cans Plasticbox & 99 & 94 & 99 & 96 & 98 & 98 & \textbf{100} & \textbf{99} & 94 & 84 & 98 & 94 \\
Place Container Plate & 98 & 97 & 96 & \textbf{100} & \textbf{99} & 98 & \textbf{99} & 97 & \textbf{99} & 95 & 98 & 99 \\
Place Dual Shoes & 91 & 88 & \textbf{94} & 88 & 93 & \textbf{89} & \textbf{94} & \textbf{89} & 75 & 75 & 93 & 87 \\
Place Empty Cup & \textbf{100} & \textbf{100} & \textbf{100} & \textbf{100} & \textbf{100} & \textbf{100} & \textbf{100} & \textbf{100} & \textbf{100} & 99 & 99 & 98 \\
Place Fan & 98 & 93 & 96 & \textbf{96} & \textbf{99} & \textbf{96} & \textbf{99} & 93 & 87 & 85 & 91 & 87 \\
Place Mouse Pad & 86 & 94 & 83 & 89 & \textbf{96} & 91 & 93 & \textbf{96} & 60 & 39 & 66 & 68 \\
Place Object Basket & 80 & 80 & 89 & \textbf{88} & 86 & 81 & \textbf{91} & \textbf{88} & 80 & 76 & 81 & 87 \\
Place Object Scale & 92 & 97 & 90 & 97 & \textbf{96} & \textbf{99} & \textbf{96} & 95 & 86 & 80 & 88 & 85 \\
Place Object Stand & 95 & 96 & 90 & 94 & 92 & \textbf{98} & \textbf{99} & 96 & 91 & 85 & 98 & 97 \\
Place Phone Stand & \textbf{100} & 96 & 97 & 99 & \textbf{100} & \textbf{100} & 97 & 97 & 81 & 81 & 87 & 86 \\
Place Shoe & 97 & 98 & 96 & \textbf{99} & 95 & 97 & 98 & 98 & 92 & 93 & \textbf{99} & 97 \\
Press Stapler & 90 & 92 & 90 & 97 & 52 & 50 & 85 & 82 & 87 & 83 & \textbf{93} & \textbf{98} \\
Put Bottles Dustbin & 94 & 91 & \textbf{95} & 90 & 93 & \textbf{95} & 87 & 91 & 84 & 79 & 81 & 79 \\
Put Object Cabinet & 87 & 89 & 94 & 89 & \textbf{95} & \textbf{90} & 85 & 87 & 80 & 79 & 88 & 71 \\
Rotate QRcode & 94 & 87 & 93 & 89 & 91 & \textbf{92} & \textbf{96} & 91 & 89 & 87 & 89 & 73 \\
Scan Object & 91 & 90 & 89 & \textbf{92} & 92 & \textbf{92} & \textbf{96} & 91 & 72 & 65 & 67 & 66 \\
Shake Bottle & \textbf{100} & \textbf{100} & \textbf{100} & \textbf{100} & \textbf{100} & \textbf{100} & \textbf{100} & 97 & 99 & 97 & \textbf{100} & 97 \\
Shake Bottle Horizontally & \textbf{100} & \textbf{100} & \textbf{100} & \textbf{100} & \textbf{100} & \textbf{100} & \textbf{100} & 99 & 99 & 99 & \textbf{100} & 98 \\
Stack Blocks Three & 98 & 96 & 95 & 97 & 98 & 97 & \textbf{99} & \textbf{98} & 91 & 76 & 91 & 95 \\
Stack Blocks Two & \textbf{100} & \textbf{100} & \textbf{100} & \textbf{100} & \textbf{100} & \textbf{100} & \textbf{100} & 98 & 97 & \textbf{100} & \textbf{100} & 98 \\
Stack Bowls Three & 83 & 78 & 80 & 81 & 84 & 86 & \textbf{86} & 83 & 77 & 71 & 79 & \textbf{87} \\
Stack Bowls Two & \textbf{98} & 96 & 92 & \textbf{98} & 97 & 95 & 94 & \textbf{98} & 95 & 96 & \textbf{98} & \textbf{98} \\
Stamp Seal & 90 & 92 & 90 & 94 & \textbf{96} & \textbf{99} & \textbf{96} & 97 & 79 & 55 & 93 & 92 \\
Turn Switch & 75 & 76 & 61 & 59 & 73 & 72 & 44 & 45 & 62 & 54 & \textbf{84} & \textbf{78} \\
\midrule
\textbf{Average} & 92.78 & \textbf{92.26} & 91.88 & 91.78 & 90.84 & 90.32 & \textbf{92.90} & 91.50 & 82.74 & 76.76 & 88.66 & 87.02 \\
\bottomrule
\end{tabular}
}
\end{table*}

\end{document}